\documentstyle[a4,12pt]{article}

\newcommand{\psr}[2]
     {\mbox{\it #1 $\rightarrow$ #2}}

\begin{document}
\title{Learning unification-based grammars
	using the Spoken English Corpus}
\author{{\bf Miles Osborne and Derek Bridge} \\
Department of Computer Science,
University of York,
Heslington, \\
York YO1 5DD,
U.\ K. \\
$\{$miles,dgb$\}$@minster.york.ac.uk }
\maketitle

\begin{abstract}
This paper describes a grammar learning system that combines model-based
and data-driven learning within a single framework.  Our results from
learning grammars using the Spoken English Corpus (SEC) suggest that
 combined model-based and data-driven learning  can  produce
a more plausible grammar than is the case when using either learning
style in isolation.

\end{abstract}
\section{Introduction}

In this paper, we present some results of our grammar learning system
acquiring unification-based grammars using the Spoken
English Corpus (SEC).  The SEC is a collection of monologues for public
broadcast and  is small ({\it circa} 50,000 words)  in comparison to other
corpora, such as the Lancaster-Oslo-Bergen
Corpus \cite{Joha78}, but sufficiently
large to demonstrate the capabilities of the learning system.  Furthermore,
the SEC is tagged and parsed, thus side-stepping the problems of constructing
a suitable lexicon and of creating an evaluation corpus to determine the
plausibility of the learnt grammars.

In contrast to other researchers (for example
\cite{Bril92,Gars87,Bake79,Lari90,Vanl87}), we try to learn competence grammars
and not performance grammars.  We also try to learn grammars that
assign linguistically plausible  parses to sentences.  Learning competence
grammars that assign plausible parses is achieved by combining model-based
and data-driven learning within a  single
framework \cite{Osbo93a,Osbo93b}.
The system is implemented to make use of
 the Grammar Development Environment (GDE) \cite{Carr88}
and it augments the GDE with
3300 lines of  Common Lisp.

Our aim in this paper is to show that combining both learning styles
produces a  grammar that assigns more plausible parses than is
then case for grammars learnt using either learning style in isolation.
Plausibility is important in Natural Language Processing as it is
very rare that applications need just to determine if a sentence is
grammatical: applications need also to determine the internal structure of
sentences (a plausible parse). A  grammar that
assigned  plausible parses  is  therefore preferable over one that
did not assign plausible parses.

The structure of this paper is as follows.  Section 2 gives an overview of
the combined model-based and data-driven learner.  Section 3 then
describes the method used to generate the results, which are then presented
in section 4.  Section 5 discusses these results and points the way
forward.

\section{System  overview}

\subsection{Architecture}

We assume that the system
has some initial grammar fragment, G,
from the outset. Presented with an input string, W, an attempt is made to parse
W using G. If this fails, the learning system is invoked. Learning takes
place through the interleaved operation of a parse completion process and
a parse rejection process.

In the parse completion process, the learning system
tries to generate rules that, had they been members of G, would have enabled
a derivation sequence for W to be found. This is done by trying to extend
incomplete derivations using what we call {\it super rules}. Super rules
are  the following unification-based grammar rules:
\begin{center}
\begin{tabular}{ll} \\
$[~]\rightarrow[~]~[~]$ & (binary) \\
$[~]\rightarrow[~]$ & (unary)
\end{tabular}
\end{center}
The binary rule says (roughly) that any category rewrites as any two other
 categories, and the unary rule says (roughly) that any category rewrites as
any
other category.
The categories in unification grammars are expressed by sets
of feature-value pairs; as the three categories in the binary
super rule and two categories in the unary super rule  specify no
values for any of the grammar's features, these rules are the most general
(or vacuous) binary and unary rules possible. These  rules thus enable
constituents found in an incomplete analysis of W to be formed into a larger
constituent. In unifying with these  constituents, the
categories on the right-hand side of the super
rules become partially instantiated with
feature-value pairs.  Hence, these rules ensure that at least one derivation
sequence will be found for W.

Many instantiations of the super rules may be produced by the parse completion
process  described above. Linguistically implausible
instantiations must be rejected and we interleave this rejection process with
the parse completion process.  Rejection of rules is carried out by the
model-driven and data-driven learning processes described below.  Note that
both of these processes are modular in design, and it would be straightforward
to add other constraints, such as lexical co-occurrence
statistics or a theory of textuality,  to help select correct analyses.

If all  instantiation are rejected, then the input string W is deemed
ungrammatical. Otherwise, surviving instantiations of the super rules
used to create the parse for W
are regarded as being linguistically plausible and may be added
to G for future use.

\subsection{Model-driven learning}

A grammatical {\it model} is a high-level theory of syntax.
In principle, if the model is complete, an `object' grammar could be produced
by computing the `deductive closure' of the model (e.g. a
`meta'-rule can be applied to those
`object' rules that account for active sentences to
produce `object' rules for passive sentences).
An example of purely model-based language learning
is given by Berwick \cite{Berw85}.  More usually, though, the model
is incomplete and this leads us to give it a different r{\^o}le in our
architecture.

Our model currently consists  of GPSG Linear Precedence (LP)
rules \cite{gkps85},
semantic types \cite{Casa88}, a Head Feature Convention \cite{gkps85} and
X-bar syntax \cite{Jack77}.

\begin{itemize}
\item {\it LP rules} are restrictions upon {\it local trees}.  A local
tree is a (sub)tree of depth one.  An example of an LP rule might
be \cite[p.50]{gkps85}:
\begin{eqnarray*}
[\mbox{SUBCAT}] \prec~ \sim [\mbox{SUBCAT}]
\end{eqnarray*}
This rule should be read as `if the SUBCAT feature is instantiated (in a
category of a local tree) then the SUBCAT feature of the linearly preceding
category should not be instantiated'.   The SUBCAT feature is used to
help indicate minor lexical categories, and so this rule states that
verbs will be initial in VPs, determiners will be  initial in NPs, and so on.
In our learning system, any putative rule that violates an LP rule is rejected.
\item We construct our syntax and semantics in tandem, adhering to the
{\it principle
of compositionality}, and pair a semantic rule to each
syntactic rule \cite{Dowt81}.  Our semantics uses the typed $\lambda$-calculus
with extensional typing.
For example, the syntactic rule:
\begin{center}
\psr{S}{NP~VP}
\end{center}
is paired with the following semantic rule:
\begin{center}
{\bf VP(NP)}
\end{center}
which should be read as `the functor {\bf VP} takes the
argument {\bf NP}'\footnote{Syntactic categories are written in a normal font
and semantic functors and
arguments
are written  in a {\bf bold} font.}.   The functor {\bf VP}
is of type\footnote{The exact details of these types is not important to
understanding the thrust of this section and so they are not given any detailed
justification.}:
\begin{eqnarray*}
<<<e, t> ,t > , t >
\end{eqnarray*}
and the argument {\bf NP} is of type:
\begin{eqnarray*}
<< e, t > ,t>
\end{eqnarray*}
The result of composing {\bf VP(NP)} has the type:
\begin{eqnarray*}
t
\end{eqnarray*}
For many newly-learnt rules, we are able to check whether the semantic
types of the categories
can be composed. If they cannot, then the
syntactic rule can be rejected.  For example, the syntactic rule:
\begin{center}
\psr{VP}{VP~VP}
\end{center}
has the semantic rule {\bf VP(VP)}, which is ill-formed because the
type
\begin{center}
$<<<e, t>, t>, t>$
\end{center}
 cannot be composed with itself.
\item Head Feature Conventions  (HFCs) help instantiate the mother
of a local tree with respect to immediately dominated daughters.  For example,
the verb phrase dominating a third person verb is itself third person.
\item X-bar syntax specifies a restriction upon the space
of possible grammar rules.  Roughly speaking, the RHS of a rule contains
a distinguished category called the {\it head} that characterises the rule.
The LHS of the rule is then a {\it projection} of the head.  Projecting
the head category results in a phrasal category of the same syntactic
class as that of the head.  For example,
the rule \psr{NP}{Det~N1} has a nominal head and a NP projection.
\end{itemize}
Model-based learning consists of filtering out instantiations of the super
rules that violate any aspect of the model, or refining instantiation of a
super rule such that they comply with some
aspect of the model.  LP rules and semantic types filter
instantiations, whilst
the Head Feature Convention and X-bar syntax refine
instantiations.

\subsection{Data-driven learning}

Our data-driven component can prefer learnt rules that are `similar' to
rules previously seen by the parser.  For this to work at all well, the
system will need some prior training
using a (pre-)training
corpus.
This  can then be used to score instantiations of the super rules.
(Note that the training set is initially  the (pre-)training corpus but
is updated as the system encounters more texts.)

The learner is trained  by
recording the frequencies of  mother-daughter pairs  (MDPs)
found in parses of sentences taken from the (pre-)training
corpus \cite{Leec91}.
For example, the tree (S (NP Sam) (VP (V laughs)))
has the following MDPs:
\begin{center}
\begin{tabular}{l}
$<$S,NP$>$ \\
$<$S, VP$>$ \\
$<$VP,V$>$
\end{tabular}
\end{center}
The frequencies of  MDPs in the parse trees previously assigned
to sentences of the training   corpus are noted.
 From these frequencies, the score  of each distinct MDP can be
computed: if pair $<$A, B$>$ occurs with frequency
$n$ out of a total number of $N$ MDPs, then the MDP's score, $f$,
is:
\begin{eqnarray*}
f(<A, B>) = n / N
\end{eqnarray*}

The  set of MDP frequencies is computed in advance
of using our system for learning. During learning, after parse completion
by the super rules, local trees in completed parses can be scored. The
score is computed recursively, as follows:
\begin{itemize}
\item For local trees of the form (A (B C)) whose daughters
are leaves,
the score of the local tree is:
\begin{eqnarray*}
score(A) = gm(f(<A,B>), \\ \nonumber f(<A,C>))
\end{eqnarray*}
where $gm$ is the geometric mean. We take the geometric mean, rather than the
product, to avoid penalising local trees that have more daughters over
local trees that have fewer daughters \cite{Mage91}.
\item For interior trees of the form (B (C D)),
the score of the local tree  is:
\begin{eqnarray*}
score(B) =   gm(score(C) \times f(<B,C>), \\   score(D) \times f(<B,D>))
\end{eqnarray*}
\end{itemize}
(This does leave the problem of dealing with MDPs
that arise in completed parses but which
did not arise in the training corpus. These can be given a low score. Giving
them
a score ensures that all trees can be scored, and thus the data-driven
learner is `complete', i.e. it can always make a decision.)

After scoring, instantiations of the super rule that
have daughters whose scores exceed some threshold can be accepted. Other
instantiations can be rejected.  The higher the threshold, the fewer the
number of rules accepted\footnote{We have not investigated the effect of
varying the threshold.  Clearly, this would be  interesting  future work.}

The approach we have described is a generalisation of the work of
Leech, who uses a simple phrase structure grammar, whereas we
use a unification-based grammar \cite{Leec87}.

\section{Method}

We predicted that the plausibility of grammars learnt using both model-based
and data-driven learning would be better than the plausibility of grammars
obtained by using either learning style in isolation.  Plausibility is
determined as how `close', for the
same sentence,  a test parse is to a benchmark parse.  The following algorithm
defines closeness between the test tree ({\it T}) and the benchmark
tree ({\it B}):
\begin{center}
\begin{itemize}
\item Each tree is normalised to use the same labelling scheme.
\item The list $L_T$ is  a preorder walk of {\it T}
and the list $L_B$ is a preorder walk of {\it B}.
\item Construct the set of lists $\alpha$  as follows.  Find
$\beta$, the longest list in both $L_T$  and $L_B$ and add $\beta$ to
$\alpha$.  Remove $\beta$ from $L_T$.  Repeat removing lists
until either $L_T$ is  the empty list or no list can be found that is
both in  $L_T$  and $L_B$.
\item Closeness is then the arithmetic mean of the list lengths of
$\alpha$ divided by the list length of $L_B$ and the nearer this figure
is to unity, the better the match.  A figure of 0 indicates no match at all.
\end{itemize}
\end{center}
To test the prediction, the following steps were taken:
\begin{center}
\begin{itemize}
\item Three disjoint sets of sentences were arbitrarily selected from
the SEC.  These were  {\it pretrain} (less than 20  sentences),
{\it train} (60 sentences) and
{\it test} (60 sentences).
\item A grammar, G,  was used as the initial grammar.  This was manually
constructed and consisted of  97 unification-based rules with a
terminal set
of the CLAWS2 tagset \cite{Blac93}.
\item The Model was configured to consist of  4 LP rules, 32 semantic types,
and a Head Feature Convention.
\item {\it Pretrain} was used to provide an initial estimate of grammaticality
for the data-driven learner.
\item {\it Train} was then processed using interleaved parsing and
learning with the following configurations of the learner:
\begin{center}
\begin{tabular}{|l|l|} \hline
Configuration & Grammar produced \\ \hline
(A) No learning & G \\
(B) Data-driven learning only & G1 \\
(C) Model-based learning only & G2 \\
(D) Both learning styles together & G3 \\ \hline
\end{tabular}
\end{center}
\item  {\it Test} was then parsed, without
learning,  using each of these grammars and the number of sentences
successfully parsed was recorded.
\item The set of sentences {\it plausible} was created as being 15
sentences in {\it test} that could be generated by grammars G1, G2
and G3.  {\it Plausible} contained no sentence that could be generated
by grammar G and hence guaranteed that each sentence needed at least one
learnt
rule in order to be generated.  As a yardstick, 15 other sentences
({\it yardstick}) that could be generated using G were
selected from {\it test}.
\item {\it Plausible} was then parsed using  grammars G1, G2 and G3 and the
first 10 parses produced for each sentence was sampled.  Out of these 10
parses, the score of the most plausible parse was noted.
\item {\it Yardstick} was parsed using grammar G and the same process
was carried out to derive 10 plausibility scores.
\end{itemize}
\end{center}

Note that X-bar syntax is such a vital aspect of acquiring plausible grammars
that it is not optional and hence all configurations use this aspect
of the model.  Configuration A is the base case for comparison with the
other configurations.

Learning grammars in the manner outlined previously is computationally
intractable.  For example, using the binary super rule may lead to
a number of parses equal (at least) to the Catalan series with respect to
sentence length.  This is because as a worst case,
the binary super rule will create all
possible binary branching parses for some sentence \cite{Chur82}.  In
order to generate
results therefore, steps were taken to place resource bounds upon the learning
process.  These bounds were to halt when n parses or m edges had been
generated (n=1, m=3000) for some sentence.  Increasing n leads to more
ambiguous attachments being learnt.  The motivation for the m limit follows
from Magerman and Weir who suggest that  large numbers of
edges being generated might correlate with
ungrammaticality \cite{Mage92}.  In effect,
the parser spends a lot of time searching unsuccessfully for a parse and
this is reflected in the large number of edges generated.  The other
constraint upon the system was that we only used the binary super rule during
interleaved parsing and learning.  This is because use of the unary rule
greatly increases the search space that needs to be explored.  The
effect of only learning
binary rules, however,  will be to decrease the plausibility
of the parses produced.

\section{Results}
 In the following table, showing some characteristics of the various grammars,
the size column is the number of rules in the  grammar, coverage
the percentage
of sentences  in {\it test} generated by each grammar,
and plausibility is the arithmetic
mean of the closeness scores  of {\it yardstick} using G and
{\it plausible} with G1, G2 and G3.
\begin{center}
\begin{tabular}{|l|l|l|l|}  \hline
Configuration & Size & Coverage  & Plausibility \\ \hline
A & 97 & 26.7  & 0.103   \\
B & 129 & 75.0& 0.086   \\
C & 128 &65.0  & 0.095    \\
D & 129 & 75.0  & 0.098    \\  \hline
\end{tabular}
\end{center}
\section{Discussion}
 Extending the initial grammar G using learning reduces G's undergeneration
considerably.  As predicted, combining model-based and data-driven learning
produces a grammar that assigns more plausible parses than do grammars
learnt using either approach in isolation.  Learnt grammars are less
plausible than the original manually constructed grammar.  The low score
given to  grammar plausibility is due to difficulties in matching
the fine-grained, steep parses produced by the unification-based grammar
with the coarse-grained, shallow parses that were manually constructed
for the SEC sentences.  The uneven quality of the SEC parses does not
help in plausibility determination.  However, the plausibility results
are encouraging and suggest that using both learning styles together is a
viable way of allowing formal grammars to be used for corpus parsing.

Future work will evaluate how much the learnt grammars overgenerate.  We
also intend investigating other constraints upon grammaticality, such as
Government and Binding Theory \cite{Chom81}, punctuation \cite{Numb90}, or
textuality \cite{Hall76,Beau81}.  Furthermore, we intend to consider using
a lexically-based formalism in place of the definite clause grammar formalism
currently used.

\section{Acknowledgements}
We would like to thank Eric Atwell (Leeds University) for allowing access
to the SEC, the anonymous referee for providing comments upon this paper,
 and Ted Briscoe (Cambridge University) for supplying
the grammar G.  The first author is supported by a Science and Engineering
Research Council grant.

\end{document}